\documentclass[a4paper,twoside,10pt]{article}

\usepackage{epsfig}
\usepackage{subcaption}
\usepackage{calc}
\usepackage{amssymb}
\usepackage{amstext}
\usepackage{amsmath}
\usepackage{amsthm}
\usepackage{multicol}
\usepackage{pslatex}
\usepackage{apalike}

\usepackage{times}
\usepackage{graphicx}
\usepackage{pgfplots}
\pgfplotsset{compat=1.6}
\usepackage{bm}
\usepackage{tikz}
\usepackage{import}
\usepackage[super]{nth}

\usetikzlibrary{positioning}
\usetikzlibrary{arrows.meta}

\usepackage{SCITEPRESS}     % Please add other packages that you may need BEFORE the SCITEPRESS.sty package.

\newcommand{\etal}{\textit{et al}. }
\newcommand{\ie}{\textit{i}.\textit{e}. }
\newcommand{\eg}{\textit{e}.\textit{g}. }

\begin{document}

\title{A Human Ear Reconstruction Autoencoder}

\author{\authorname{Hao Sun\sup{1}\orcidAuthor{0000-0003-2062-127X}, Nick Pears\sup{1}\orcidAuthor{0000-0001-9513-5634} and Hang Dai\sup{2}\orcidAuthor{0000-0002-7609-0124}}
\affiliation{\sup{1}Department of Computer Science, University of York, York, UK}
\affiliation{\sup{2}Mohamed bin Zayed University of Artificial Intelligence, Abu Dhabi, UAE}
\email{\{hs1145, nick.pears\}@york.ac.uk, hang.dai@mbzuai.ac.ae}
}


\abstract{\small{The ear, as an important part of the human head, has received much less attention compared to the human face in the area of computer vision. Inspired by previous work on monocular 3D face reconstruction using an autoencoder structure to achieve self-supervised learning, we aim to utilise such a framework to tackle the 3D ear reconstruction task, where more subtle and difficult curves and features are present on the 2D ear input images. Our Human Ear Reconstruction Autoencoder (HERA) system predicts 3D ear poses and shape parameters for 3D ear meshes, without any supervision to these parameters. To make our approach cover the variance for in-the-wild images, even grayscale images, we propose an in-the-wild ear colour model. The constructed end-to-end self-supervised model is then evaluated both with 2D landmark localisation performance and the appearance of the reconstructed 3D ears.}}

\onecolumn \maketitle \normalsize \setcounter{footnote}{0} \vfill

%% Body text begins.

\section{\uppercase{Introduction}}

\noindent Three-dimensional (3D) face modelling and 3D face reconstruction from monocular images have drawn increasing attention over the last few years. Especially with deep learning methods, 3D face reconstruction models are empowered to have more complexity and better feature extraction ability. However, as an important part of the human head, the human ear has significantly less relative literature published. Being able to reconstruct 3D ears implies establishing the dense correspondence between 3D ear vertices and 2D ear image pixels, thus enabling further applications such as 3D/2D ear landmark localisation and ear recognition \cite{zhou2017deformable,emervsivc2017ear,emervsivc2019unconstrained}. Furthermore, it can be a vital part of constructing a 3D model of the full human head \cite{dai20203d,ploumpis2020towards}.

\begin{figure}[ht]
    \centering
    \resizebox{\columnwidth}{!}{
        \begin{tikzpicture}
            \node (lm_semantics) [inner sep=0pt, anchor=south east, label=below:(1)] at (0,0)
            {\includegraphics[width=4cm]{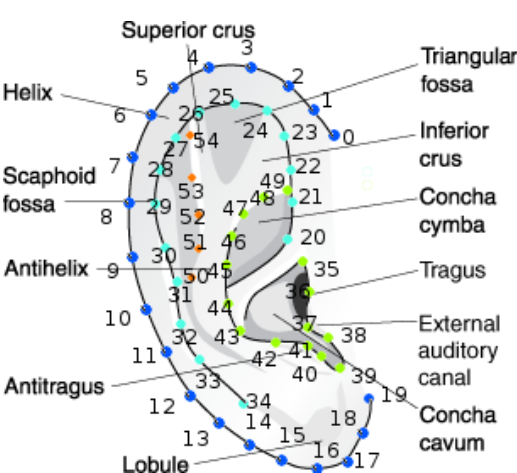}};
            
            \node (render) [right = 0.2cm of lm_semantics, inner sep=0pt, label=below:(2)]
            {\includegraphics[width=4cm]{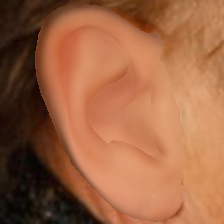}};
            
            \node (landmarks) [right = 0.2cm of render, inner sep=0pt, label=below:(3)]
            {\includegraphics[width=4cm]{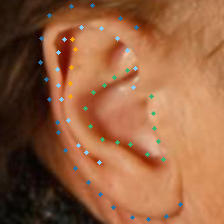}};
        \end{tikzpicture}
    }
    \caption{(1) $55$ landmarks and their semantics from ITWE-A dataset \cite{zhou2017deformable} (2) Rendered densely corresponded coloured 3D ear mesh projected onto the original image (3) Original image marked with predicted landmarks.}
    \label{fig:intro}
\end{figure}

Most modern approaches for 3D face or 3D ear reconstruction from monocular images fall into three categories: generation based, regression based and the combination of both \cite{tewari2017mofa}. Generation based methods require a parametric model for the 3D object and 3D landmarks to optimise a set of parameters for optimal alignment between projected 3D models and 2D landmarks. For 3D ear reconstructions, two approaches can be found in literature \cite{dai2018data,zhou2017deformable}. Regression-based methods usually utilise neural networks to regress a parametric model's parameters directly, as proposed by \cite{richardson20163d} for 3D face reconstruction. Generation-based methods are often more computationally costly, due to their non-convex optimisation criteria and the requirement for landmarks. Regression-based methods require ground truth parameters to be provided, which is only accessible when using synthetic data \cite{richardson20163d}. Otherwise other 3D reconstruction algorithms are required to obtain ground truth parameters beforehand \cite{zhu2017face}. Therefore, Tewari \etal proposed a self-supervised 3D face reconstruction method named \emph{Model-based Face Autoencoder}  (MoFA) that combines both generation and regression based methods. This aims to mitigate the negative aspects of the two categories of method, by using an autoencoder composed of a regression-based encoder and a generation-based decoder \cite{tewari2017mofa}. However, there are no regression-based or autoencoder structured approaches for 3D ear reconstruction in the literature. Whether this self-supervised autoencoder approach can tackle the complexity of the ear structure remains an open question that we address here.

The core idea of the self-supervised learning approach is to synthesise similar colour images from original colour input images in a differentiable manner. For such an approach, a parametric ear model is needed. Dai \etal propose a 3D Morphable Model (3DMM) of the ear, named the York Ear Model (YEM). Its 3D ear mesh has $7111$ vertex coordinates, so $21333$ vertex parameters, reduced to $499$ shape parameters using PCA. However, to enable self-supervised learning, the 3D ear meshes require colour/texture, which is not included in the YEM model.

In this context, we present a Human Ear Reconstruction Autoencoder (HERA) system, with the following contributions:
\begin{itemize}
    \item A 3D ear reconstruction method that is completely trained unsupervised using in-the-wild monocular ear colour 2D images.
    \item An in-the-wild ear colour model that colours the 3D ear mesh to minimise its difference with the 2D ear image in appearance.
    \item Evaluations that demonstrate that the proposed model is able to predict a densely corresponded coloured 3D ear mesh (\eg Figure \ref{fig:intro} (2)) and 2D landmarks (\eg Figure \ref{fig:intro} (3)).
\end{itemize}

\begin{figure*}[t]
    \centering
    \resizebox{\textwidth}{!}{
        \subimport{}{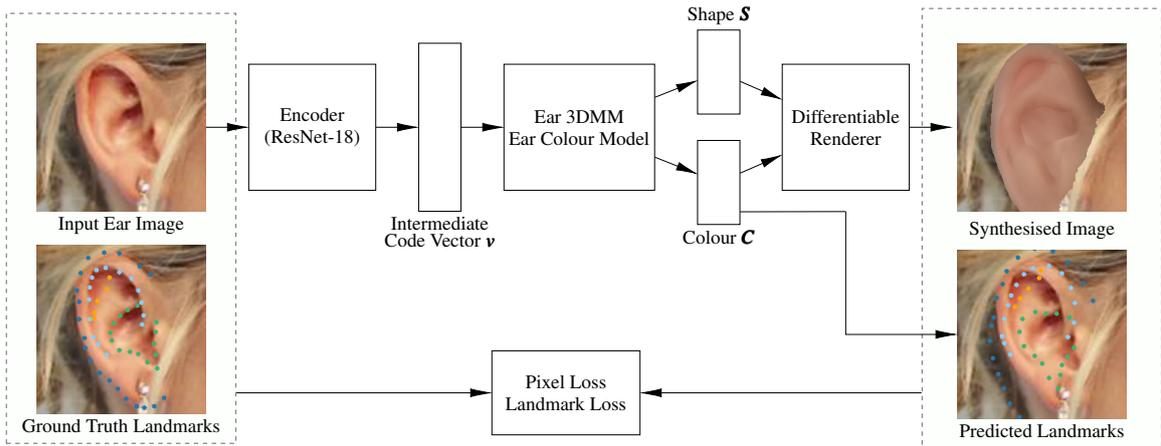}
    }
    \caption{Overview of the autoencoder architecture}
    \label{fig:arch}

\end{figure*}

\section{\uppercase{Related Work}}

\noindent In this section, we discuss a range of 3D face reconstruction methods that utilise an autoencoder structure to achieve self-supervised learning. The method this paper proposes obtains 3D ear shapes by employ a strong prior provided by an ear 3DMM, thus the two existing 3D parametric ear models will be discussed. Finally, two methods that evaluate their methods using normalised landmark error are discussed, since we evaluate landmark prediction accuracy on the same dataset, using the same metric.

\subsection{Self-supervised Learning for 3D Dense Face Reconstruction}

\noindent The self-supervised learning approach to 3D face reconstruction builds an end-to-end differentiable pipeline that takes the original colour images as input, predicts and reconstructs the 3D face mesh, then uses a differentiable renderer to reconstruct colour images as output. The goal of such a self-supervised learning approach is to minimise the difference between input colour images and output colour images. Several novel 3D face reconstruction approaches have recently been proposed. Improvements include using a face recognition network to contribute to a loss function, using a Generative Adversarial Network (GAN) for texture generation \cite{gecer2019ganfit} and replacing the linear 3DMM structure with a non-linear 3DMM \cite{tran2018nonlinear}. The aim of all of those approaches is to achieve better performance more intuitively, particularly in terms of minimising the appearance difference between generated output images and real input images.

\subsection{In-the-wild Ear Image Dataset}

\noindent There are numerous in-the-wild ear image datasets built for various purposes, here we focus on Collection A from the \emph{In-the-wild Ear Database} (ITWE-A) since it has $55$ manually-marked landmarks. All the landmarks have semantic meaning, as shown in Figure \ref{fig:intro} (1). This dataset contains $500$ images in its training set and $105$ images in its test set, where each image is captured in-the-wild and contains a clear ear. The dataset has a large variation in ear colours, as is the nature of in-the-wild images, and it even contains several grayscale images. Traditional 3DMM colour models, such as that of the Basel Face Model 09 (BFM09) \cite{blanz1999morphable}, often fail to generate a highly-similar appearance to the input. However, the in-the-wild ear colour model proposed here, can cover such colour variance, since it models directly from the in-the-wild images themselves.

\subsection{Parametric Ear Models}

\noindent Zhou and Zaferiou build their parametric ear model using an Active Appearance Model (AAM), which is a linear model that aims to model the 2D ear's shape and colour simultaneously\cite{cootes1998active}. A 3D Morphable Model (3DMM) is a closely-related model that models objects' shapes and colours in 3D instead of 2D. Blanz and Vetter first proposed a 3D Morphable Model (3DMM) for human faces \cite{blanz1999morphable}, which builds a linear system that allows different 3D face meshes to be described by $199$ shape parameters. Similarly, Dai \etal \cite{dai2018data} proposed a 3D morphable model for the human ear named the York Ear Model (YEM), also based on a linear system, but with $499$ parameters. Here, we utilise this ear 3DMM for its strong 3D ear shape prior. Meanwhile, the reduced dimension of the parameters allows the neural network to perform a much easier regression task using $499$ shape parameters rather than $21333$ raw vertex parameters.

\subsection{2D Ear Detection}

\noindent Ear detection or localisation in 2D images aims to find the region of interest bounding the ear, from images of the human head that contain ears; for example, profile-view portraits. It is a vital preprocessing step in the 3D ear reconstruction pipeline. Object detection has been studied for decades and there exists a number of algorithms that specifically perform the 2D ear detection task. Zhou and Zaferiou \cite{zhou2017deformable} use the histogram of oriented gradients with a support vector machine (HoG+SVM) to predict a rectangular region of interest. Emer\v{s}i\v{c} \etal \cite{emervsivc2017pixel} and Bizjak \etal \cite{bizjak2019mask} propose deep learning methods to tackle the 2D ear detection task by predicting a pixel-level segmentation of the 2D ear image directly.

\subsection{2D Ear Landmark Localisation}

\noindent 2D ear landmark localisation is a task for finding specific key points on 2D ear images. It is an intuitive method of quantitative evaluation of this work where the shape and alignment of the reconstructed 3D ear mesh can be evaluated precisely. In 2D face landmark localisation, numerous approaches obtain 2D landmarks by reconstructing 3D models first \cite{zhu2017face,liu2016joint,mcdonagh2016joint}. Being able to achieve competitive results against a specialised 2D landmark predictor is necessary for the success of a 3D dense ear reconstruction algorithm. Zhou and Zaferiou's approach comes with the ITWE-A dataset and is considered as a baseline. They use Scale Invariant Feature Transform (SIFT) features and an AAM model to predict 2D landmarks \cite{zhou2017deformable}. Hansley and Segundo \cite{hansley2018employing} propose a CNN-based approach to regress 2D landmarks directly and they also evaluate on the ITWE-A dataset. Their approach proposes two CNNs that both predict the same set of landmarks but with different strengths. The first CNN has better generalisation ability for different ear poses. The resulting landmarks of the first CNN are used to normalise the ear image. The second CNN predicts improved normalised ear images based on the results of the first CNN.

\section{\uppercase{The HERA system}}

\noindent Our proposed Human Ear Reconstruction Autoencoder (HERA) system employs an autoencoder structure that takes ear images as input and generates synthetic images. Therefore, it is trained by minimising the difference between input images and the final synthesised images. An illustration of our end-to-end architecture is shown in Figure \ref{fig:arch}. The encoder is a CNN predicting intermediate code vectors that are then fed to the decoder, where coloured 3D ear meshes are reconstructed and rendered into 2D images. 

The decoder is comprised of: \emph{(1)} the YEM ear shape model and our in-the-wild ear colour model that reconstruct ear shapes and ear colours respectively; \emph{(2)} PyTorch3D \cite{ravi2020pytorch3d} that renders images with ear shapes and colours in a differentiable way. The comparison of the input and synthesised images is implemented by a combination of loss functions and regularisers. The essential loss function is a photometric loss, with an additional landmark loss that can be included for both faster convergence time and better accuracy. The whole autoencoder structure is designed to be differentiable and so can be trained in an end-to-end manner. Each part of the architecture (\ie encoder CNN, ear 3DMM, scaled orthogonal projection and loss functions) is differentiable by default, thereby using a differentiable renderer to render 3D meshes to 2D images makes the whole architecture differentiable. The core part of the decoder is described in Section \ref{sec:ear-3dmm}. The whole end-to-end trainable architecture and the necessary training methods are then described in Section \ref{sec:ear-ae}.

\subsection{Ear 3D Morphable Model Preliminaries}
\label{sec:ear-3dmm}
\noindent This section describes the 3DMM part of the decoder which comprises an ear shape model derived from the YEM, an ear colour model, and the projection model. With this 3DMM, the shape parameters $\bm{\alpha_}{s}$ can be reconstructed to an 3D ear vertex coordinate vector $\bm{S} \in \mathbb{R}^{N\times3}$ where $N$ is the number of vertices in a single 3D ear mesh. The colour parameters $\bm{\alpha_}{c}$ are then reconstructed to a vertex colour vector $\bm{C} \in \mathbb{R}^{N\times3}$ to colour each vertex. The pose parameters $\bm{p}$ are used in the projection model that aligns 3D ear meshes with 2D ears' pixels.

\subsubsection{Ear Shape Model}

\noindent We employ YEM model \cite{dai2018data}, which supplies the geometric information necessary for reconstruction. It is constructed using PCA from $500$ 3D ear meshes and thus provides a strong statistical prior. The 3D ear vertex coordinate vector (\ie 3D ear shape) $\bm{S}$ is reconstructed from shape parameter vector $\bm{\alpha}_S$ by:
\begin{equation}
\label{eq:shape-3dmm}
    \bm{S} = \hat{S}\left(\alpha_{s}\right) = \Bar{\bm{S}} + \bm{U}_s\bm{\beta}_{s},
\end{equation}
where $\Bar{\bm{S}} \in \mathbb{R}^{3N}$ is the mean ear shape, $\bm{U}_s \in \mathbb{R}^{3N \times 499}$ is the ear shape variation components and the resulting matrix is rearranged into a $N \times 3$ matrix, where each row represents a vertex coordinate in 3D space.

The projection model employed is the scaled orthogonal projection ($SOP$) projecting 3D shape to 2D. Given the 3D ear shape $\bm{S}$ from Equation \ref{eq:shape-3dmm}, the projection function, $\hat{V}$, is defined as:
\begin{equation}
\label{eq:sop}
    \bm{V} = \hat{V}\left( \bm{S}, \bm{p} \right) = f \bm{P}_{o} \hat{R}\left(\bm{r}\right) \bm{S} + \bm{T},
\end{equation}
where $\bm{P}_{o} = \begin{bmatrix} 1 & 0 & 0 \\ 0 & 1 & 0 \end{bmatrix}$ is the orthogonal projection matrix, $\bm{V} \in \mathbb{R}^{N \times 2}$ are the projected 2D ear vertices and $\hat{R}\left(\bm{r}\right)$ is the function that returns the rotation matrix. Since scaled-orthogonal projection is used, $\bm{V}$ provides sufficient geometric information for the differentiable renderer and no additional camera parameters are needed.

In addition, 2D landmarks can be extracted from the projected vertices $\bm{V}$ by manually selecting $55$ semantically corresponding vertices. Thus we can define a vector of 2D landmarks of a projected ear shape $\bm{V}$ as:
\begin{equation}
\label{eq:lm-selecting}
    \bm{X}_{i} = \bm{V}\left( \bm{L} \right),
\end{equation}
where $\bm{X}_{i} \in \mathbb{R}^{55 \times 2}$ are the landmark's x and y coordinates indexed by $\bm{L}$ in the projected ear vertices $\bm{V}$.

\subsubsection{In-the-wild Ear Colour Model}

\noindent The YEM model contains an ear shape model only. However, the decoder in our architecture requires the 3D ear meshes to be coloured to generate plausible synthetic ear images. To solve this problem, we build an in-the-wild ear colour model using PCA whitening.

Firstly, for each ear image from of the $500$ images from the training set of the ITWE-A dataset, a set of whitened ear shape model parameters $\bm{\alpha}_{s}$ and ear pose $\bm{p}$ is fitted using a non-linear optimiser to minimise 2D landmark distances. Using the reconstruction Equations \ref{eq:pca-whitening} $\sim$ \ref{eq:lm-selecting}, the optimisation criteria $E_{0}$ can be formed as follow:
\begin{gather}
    \hat{X}\left( \bm{\alpha}_{s}, \bm{p} \right) =  \hat{V}\left( \hat{S}\left( \hat{\alpha}\left(\bm{\alpha}_{s} \right) \right), \bm{p} \right), \\
    E_{0}\left( \bm{\alpha}_{s}, \bm{p}, \bm{X}_{gt} \right) = \frac{1}{N_{L}}\left\| \left( \hat{X}\left( \bm{\alpha}_{s}, \bm{p} \right) \right)\left( \bm{L} \right) - \bm{X}_{gt}\right\|_{2},
\end{gather}
where $\hat{X}$ is the whole reconstruction and projection function, $N_{L} = 55$ is a constant representing the number of landmarks and $\bm{X}_{gt} \in \mathbb{R}^{55 \times 2}$  is the ground truth 2D landmarks provided by the ITWE-A dataset.

\begin{figure}
    \centering
    \resizebox{\columnwidth}{!}{
        \subimport{}{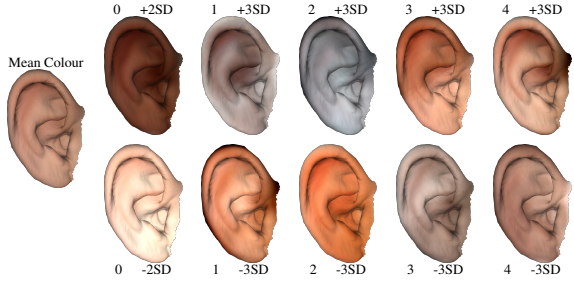}
    }
    \caption{In-the-wild Ear Colour Model. The mean colour and first 5 parameters $\pm$ standard deviations (SD) are shown. The mean 3D ear mesh is used.}
    \label{fig:colour_model}

\end{figure}

After the shapes are fitted, the colour for each vertex is obtained by selecting the corresponding 2D pixel colour. This process ends up in $500$ vertex colour vectors, which can then be used to build the in-the-wild ear colour model using PCA whitening. The vertex colour vectors are parameterised by $40$ parameters and cover by $86.6\%$ of the colour variation. The reconstruction coverage rate is not proportional to the quality of the model building, since setting a moderate coverage rate can implicitly ignore some occlusions (\eg hair and ear piercings). This colour model is shown in Figure \ref{fig:colour_model}.

The reconstruction of the vertex colour vector $\bm{C}$ is:
\begin{equation}
    \bm{C} = \hat{C}\left( \bm{\alpha}_{c} \right) = \Bar{\bm{C}} + \bm{U}_{c}\bm{\alpha}_c,
\end{equation}
where $\bm{\alpha}_{c} \in \mathbb{R}^{40 \times 1}$ is the colour parameter vector. $\Bar{\bm{C}}$ is average vertex colour vector, $\bm{U}_{c}$ is vertex colour variance component matrix and both are calculated by the PCA whitening algorithm.

\subsection{Intermediate Code Vector}

\noindent The intermediate code vector
\begin{equation}
    \bm{v} = \left\{ \bm{p}, \bm{\alpha}_{s}, \bm{\alpha}_{c} \right\}
\end{equation}
connects the encoder and the decoder and has semantic meaning. Where
\begin{equation}
    \bm{p} = \left\{ \bm{r}, \bm{T}, f \right\}
\end{equation}
defines the pose of the 3D ear mesh. $\bm{r} \in \mathbb{R}^{3}$ is the azimuth, elevation and row which map to the rotation matrix through function $\hat{R}\left(\bm{r}\right) : \mathbb{R}^{3} \rightarrow \mathbb{R}^{3 \times 3}$. $\bm{T} \in \mathbb{R}^{2\times1}$ defines the translation in X-axis and Y-axis. The translation in z-axis is not necessary since scaled orthogonal projection is used. $f$ is a fraction number that defines the 3D mesh's scale. $\bm{\alpha}_{s} \in \mathbb{R}^{40 \times 1}$ are the PCA whitened shape parameters and will be recovered to the shape parameters $\bm{\beta}_{s} \in \mathbb{R}^{499 \times 1}$ and then proceeded by the YEM 3DMM. $\bm{\alpha}_{c} \in \mathbb{R}^{40 \times 1}$ are the colour parameters for the in-the-wild ear colour model built by this paper.

\subsection{PCA Whitening}

\noindent To ease the optimisation process in training, we use PCA whitening to transfer the YEM ear model parameters into the format that is more favourable for deep learning frameworks. Firstly, the variances of the parameters can differ in a very large scale from $8 \times 10^{3}$ for the most significant parameter to $5 \times 10^{-7}$ for the least important parameter. It is difficult to train a neural network to effectively regress such large variance data. Secondly, the large number of the parameters slow the neural networks' training speed and worse the optimisation process. This could be mitigated by trimming a portion of the less important parameters out. But this has potential to lose the shape and color information from the trimmed part. To overcome this, we perform PCA whitening \cite{kessy2018optimal} over the full set of parameters. PCA whitening aims to generate zero-mean parameters with reduced dimensions in unit-variance. In our experiment, YEM's original parameters $\bm{\beta}_{s}$ of $499$ dimensions are transformed to $\bm{\alpha}_{s}$ of $40$ dimensions while covering $98.1\%$ of the variance associated with the original parameters. Each original parameter vector $\bm{\beta}_{s}$ can be recovered from $\bm{\alpha}_{s}$ by:
\begin{equation}
\label{eq:pca-whitening}
    \bm{\beta}_{s} = \hat{\alpha}\left( \bm{\alpha}_{s} \right) = \bm{U}_{w}\bm{\alpha}_{s},
\end{equation}
where $\bm{U}_{w} \in \mathbb{R}^{499 \times 40}$ is a constant matrix of variation components calculated by the PCA whitening procedure. The original parameters' mean is not added since they are zero-mean already.

\subsection{Ear Autoencoder}
\label{sec:ear-ae}

\noindent We now combine the intermediate code vector and decoder components, described in previous sections, 
with the encoder, the differentiable renderer and the loss functions, to build the end-to-end autoencoder 

As illustrated in Figure \ref{fig:arch}, we build an self-supervised architecture that consists of an encoder, an intermediate code vector, the decoder components, the differentiable renderer and the loss for back-propagation. 

The encoder is an 18-layer residual network (ResNet-18) which is a CNN that performs well on regression from image data \cite{he2016deep}. We use PyTorch3D \cite{ravi2020pytorch3d} as a differentiable image renderer developed using PyTorch \cite{paszke2019pytorch}. It is a differentiable function that maps a set of vertex coordinate vector and vertex colour vector to a 2D image. The encoder $Q$ and decoder $W$ can be formed as follows:
\begin{gather}
    \bm{v}_{pred} = Q\left(\bm{I}_{in}, \bm{\theta}\right),\\
    \bm{S}_{pred}^{T}, \bm{C}_{pred} = W\left(\bm{v}_{pred}\right),\\
    \bm{I}_{pred} = Render\left(\bm{S}_{pred}^{T}, \bm{C}_{pred}\right),\\
    \bm{X}_{pred} = \bm{S}_{pred}^{T}\left(\bm{L}\right),
\end{gather}
% \begin{gather}
%     \bm{v}_{pred} = Q\left(\bm{I}_{in}, \bm{\theta}\right),\\
%     \bm{S}_{pred}, \bm{C}_{pred} = W_{r}\left(\bm{v}_{pred}\right),\\
%     \bm{I}_{pred} = W_{im}\left(\bm{S}_{pred}, \bm{C}_{pred}\right),\\
%     \bm{X}_{pred} = W_{l}\left(\bm{S}_{pred}\right),
% \end{gather}
where $\bm{I}_{in}$ is the input image and $\bm{\theta}$ are the weights of the encoder network $Q$. In the decoder $W$, the predicted 3D mesh (\ie shape with pose $\bm{S}_{pred}^{T}$ and colour $\bm{C}_{pred}$) are reconstructed from the predicted intermediate code vector $\bm{v}_{pred}$. The reconstructed 3D mesh is then fed to the differential image render for capturing the rendered image $\bm{I}_{pred}$. The $\bm{L}$ indexes the x and y coordinates of the 55 ear landmarks in the ear shape $\bm{S}$. The predicted landmarks $\bm{X}_{pred} \in \mathbb{R}^{55 \times 2}$ can be derived from the predicted ear shape by indexing the x and y coordinates of the 55 ear landmarks in the predicted 3D ear shape from $\bm{L}$. The encoder ResNet-18 is initialised using the weights pre-trained on ImageNet \cite{deng2009imagenet}. The trained encoder network can be used for the shape and color parameters regression.

\subsubsection{Loss Function}
\label{sec:loss-functions}

\noindent Our loss function follows the common design of loss functions in differentiable renderer based self-supervised 3D reconstruction approaches. The proposed loss function is a combination of four weighted losses as:
\begin{multline}
    E_{loss} = \lambda_{pix}E_{pix}\left( \bm{I}_{in} \right) + \lambda_{lm}E_{lm}\left( \bm{I}_{in}, \bm{X}_{gt}\right)
    \\ + \lambda_{reg1}E_{reg1}\left( \bm{I}_{in} \right) + \lambda_{reg2}E_{reg2}\left( \bm{I}_{in} \right),
\end{multline}
where $\lambda_{i}$ are the weights for the losses $E_{i}$.

\paragraph{Pixel Loss}
The core idea of the self-supervised architecture is that the model can generate synthetic images from input images and are compared with input images. Thus to form such comparison, the Mean Square Error (MSE) is used on all pixels:
\begin{equation}
    E_{pix}\left( \bm{I}_{in} \right) = L_{MSE}\left( Render\left(W \left( Q\left( \bm{I}_{in}, \bm{\theta} \right) \right) \right), \bm{I}_{in} \right),
\end{equation}
Where $L_{MSE}$ is a function that calculates the mean square error. A pixel mask is used to compare the rendered ear region only, since the rendered ear images have no background.

\paragraph{Landmark Loss}
The optional landmark loss is used to speed up the training process and help the network learn 3D ears with better accuracy. Zhou and Zaferiou \cite{zhou2017deformable} propose the mean normalised landmark distance error as their shape model evaluation metric. Here, we employ it as a part of the loss function. 
It can be formed as:
\begin{multline}
\label{eq:landmark-loss}
    E_{lm}\left( \bm{I}_{in}, \bm{X}_{gt}\right) = \frac{\left\| \left(W \left( Q\left( \bm{I}_{in}, \bm{\theta} \right) \right) \right) \left( \bm{L} \right) - \bm{X}_{gt} \right\|_{2}}
    {D_N\left( \bm{X}_{gt} \right)N_{L}}
\end{multline}
where $\bm{X}_{gt}$ is the ground truth landmarks and $D_N\left( \bm{X}_{gt} \right)$ is a function gets the diagonal pixel length of the ground truth landmarks' bounding box. Since this loss is optional, setting $\lambda_{lm} = 0$ can enable the whole model to be trained on 2D image data $\bm{I}_{in}$ only, making the use of very large-scale unlabelled training data possible.

\paragraph{Regularisers}
Two regularisers are used to constrain the learning process and are weighted separately. The first regulariser is the statistical plausibility regulariser.
%following the basic assumption of the PCA whitening algorithm that each parameter has zero mean. 
The regulariser is formed by:
\begin{equation}
    E_{reg1}\left( \bm{I}_{in} \right) = \sum_{j=1}^{40}\bm{\alpha}_{sj} + \sum_{j=1}^{40}\bm{\alpha}_{cj},
\end{equation}
where $\bm{\alpha}_{s}$ and $\bm{\alpha}_{c}$ are ear shape and colour parameters predicted by the encoder network. Therefore this penalises the Mahalanobis distance from the mean shape and colour.

During our experiments, we found that an additional restriction on the scale parameter $f$ has to be applied for the model to be successfully trained without landmarks. The restriction is formed by:
\begin{equation}
    E_{reg2}\left( \bm{I}_{in} \right) =
    \begin{cases}
    \left(0.5 - f\right)^{2} & \text{if } f < 0.5\\
    \left(f - 1.5\right)^{2} & \text{if } f > 1.5\\
    0 & \text{otherwise}
    \end{cases},
\end{equation}
We employed two sets of weights, $\lambda$, depending on whether or not landmark loss is used when training. 
\begin{itemize}
    \item Training with landmarks: $\lambda_{pix} = 10$, $\lambda_{lm} = 1$, $\lambda_{reg1} = 5\times10^-2$ and $\lambda_{reg2} = 0$
    \item Training without landmarks: $\lambda_{pix} = 2$, $\lambda_{lm} = 0$, $\lambda_{reg1} = 5\times10^-2$ and $\lambda_{reg2} = 100$
\end{itemize}

\subsubsection{Dataset Augmentation}

\noindent Since the ITWE-A dataset used to train our model contains only $500$ landmarked ear images, having limited variance on ear rotations, we perform data augmentation on the original dataset. An ear direction of a 2D ear image is defined by a 2D vector from one of the ear lobe landmark points to one of the ear helix landmark points. For each 2D ear image, $12$ random rotations around its central point are applied such that the angles between their ear directions and the Y-axis of the original image are uniformly distributed between $-60^{\circ}$ and $60^{\circ}$. The augmented ear image dataset contains $6,000$ images in total. With this augmentation, we find that test set landmark error drops significantly.

\section{\uppercase{Results}}

\noindent In this section, both quantitative evaluation results and qualitative evaluation results are discussed. Quantitative evaluation focuses on comparing landmark fitting accuracy with different approaches. While the qualitative evaluation focuses on evaluating the visual results of this 3D ear reconstruction algorithm. Furthermore, an ablation study is conducted to analyse the improvement that various optimisations of this work has proposed, including the PCA whitening on the YEM model parameters, the statistical plausibility regulariser and the dataset augmentation. The abbreviation Human Ear Reconstruction Autoencoder (HERA) is used to represent the final version of this work.

\subsection{Quantitative Evaluations}

\noindent The main quantitative evaluation method applied is the mean normalised landmark distance error proposed by \cite{zhou2017deformable} formed in Equation \ref{eq:landmark-loss} which also forms the landmark loss that trains our system. Projecting the 3D ear meshes' key points to 2D and comparing them with the ground truth can assess the accuracy of the 3D reconstruction. There are two approaches that predict the same set of landmarks using the same dataset in the literature , therefore comparisons can be formed. Zhou \& Zaferiou's work  \cite{zhou2017deformable} is considered as a baseline solution and Hansley \& Segundo's work \cite{hansley2018employing} is a specifically designed 2D landmark localisation algorithm that has the lowest landmark error in the literature. To interpret the landmark error, it is stated that for an acceptable prediction of landmarks, the mean normalised landmark distance error has to be below $0.1$ \cite{zhou2017deformable}. This is a dimensionless metric that is the ratio of the mean Euclidean pixel error to the diagonal length of the ear bounding box.

As this paper stated in Section \ref{sec:loss-functions},  HERA can be trained without landmarks or data augmentation in a self-supervised manner. The HERA version that uses no landmark loss during training and trains on the original $500$ ear images is named HERA-W/O-AUG-LM.

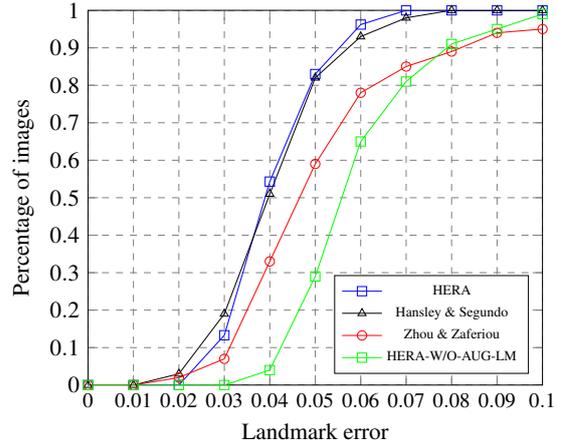
\begin{figure}
    \centering
    \resizebox{\columnwidth}{!}{
        \begin{tikzpicture}
            \begin{axis}[
                xlabel={Landmark error},
                ylabel={Percentage of images},
                xmin=0, xmax=0.1,
                ymin=0, ymax=1.0,
                xtick={0,0.01,0.02,0.03,0.04,0.05,0.06,0.07,0.08,0.09,0.10},
                ytick={0,0.1,0.2,0.3,0.4,0.5,0.6,0.7,0.8,0.9,1.0},
                x tick label style={/pgf/number format/fixed, font=\footnotesize},
                ymajorgrids=true,
                grid style={dashed,gray},
                grid=both,
                legend style={font=\tiny},
                legend pos=south east,
            ]
            
            \addplot[
                color=blue,
                mark=square,
                ]
                coordinates {
                (0.0,0)(0.01,0)(0.02,0)(0.03,0.133)(0.04,0.543)(0.05,0.83)(0.06,0.962)(0.07,1.0)(0.08,1.0)(0.09,1.0)(0.10,1.0)
                };
            \addplot[
                color=black,
                mark=triangle,
                ]
                coordinates {
                (0.0,0)(0.01,0)(0.02,0.03)(0.03,0.19)(0.04,0.51)(0.05,0.82)(0.06,0.93)(0.07,0.98)(0.08,1.0)(0.09,1.0)(0.10,1.0)
                };
            \addplot[
                color=red,
                mark=o,
                ]
                coordinates {
                (0.0,0)(0.01,0)(0.02,0.02)(0.03,0.07)(0.04,0.33)(0.05,0.59)(0.06,0.78)(0.07,0.85)(0.08,0.89)(0.09,0.94)(0.10,0.95)
                };
            \addplot[
                color=green,
                mark=square,
                ]
                coordinates {
                (0.0,0)(0.01,0)(0.02,0.00)(0.03,0.00)(0.04,0.04)(0.05,0.29)(0.06,0.65)(0.07,0.81)(0.08,0.91)(0.09,0.95)(0.10,0.99)
                };
                \legend{HERA, Hansley \& Segundo, Zhou \& Zaferiou, HERA-W/O-AUG-LM}
            \end{axis}
        \end{tikzpicture}
    }
    \caption{Cumulative error distribution curve comparison among different landmark detection algorithms and our work}
    \label{fig:quant-ced_curve}
\end{figure}

Our HERA system is now compared with Zhou \& Zaferiou's and Hansley \& Segundo's work regarding the normalised landmark error's mean, standard deviation, median and cumulative error distribution (CED) curve evaluated on the test set of ITWE-A which contains $105$ ear images. The numerical results are shown in Table \ref{tab:quant-res} and the CED curve is shown in \ref{fig:quant-ced_curve}. Additionally, the percentage of predictions that have error less than $0.1$ and $0.6$ are given in Table \ref{tab:quant-res}.

\begin{table}
    \caption{Normalised landmark distance error statistics on ITWE-A.}
    \begin{center}
        \resizebox{\columnwidth}{!}{
            \begin{tabular}{|l|c|c|c|c|}
                \hline
                Method & mean $\pm$ std & median & $\leq 0.1$ & $\leq 0.06$\\
                \hline\hline
                Zhou \& Zaferiou & $0.0522 \pm 0.024$ & $0.0453$ & $95\%$ & $78\%$ \\
                HERA & $\bm{0.0398 \pm 0.009}$ & $\bm{0.0391}$ & $\bm{100\%}$ & $\bm{96.2\%}$ \\
                HERA-W/O-AUG-LM & $0.0591 \pm 0.014$ & $0.0567$ & $99\%$ & $64.7\%$ \\
                \hline
            \end{tabular}
        }
    \end{center}
    \label{tab:quant-res}
\end{table}

From Table \ref{tab:quant-res} and Figure \ref{fig:quant-ced_curve}, it can be concluded that HERA outperforms Zhou \& Zaferiou's work by a large margin in terms of 2D landmark localisation task. When compared with Hansley \& Segundo's 2D landmark localisation work, similar results are shown. This is considered acceptable when comparing a 3D reconstruction algorithm with a 2D landmark localisation algorithm. Hansley \& Segundo's landmark localiser is comprised of two specifically designed CNNs for landmark regressions while HERA uses only one CNN to regress a richer set of information (\ie pose, 3D model's parameters and colour parameters). Regarding the threshold of $0.1$ proposed by \cite{zhou2017deformable}, both HERA and  Hansley \& Segundo's work are $100\%$ below $0.1$, and HERA trained without landmarks achieves $99\%$ below $0.1$. The CED curves show that, although HERA-W/O-AUG-LM performs worse than Zhou \& Zaferiou's work in the error region below around 0.077, our performance is better at the 0.1 error point.  In other words, HERA-W/O-AUG-LM can predict landmarks with less than $0.1$ error more consistently than the baseline.

% It takes the trained network $\sim6$ ms to predict the intermediate code vector of a single cropped ear image on an Nvidia RTX 2080.

\subsection{Qualitative Evaluations}

\begin{figure}
    \centering
    \resizebox{\columnwidth}{!}{
        \begin{tikzpicture}
            \node (0_lm) [inner sep=0pt, anchor=south east] at (0,0)
            {\includegraphics[width=4cm]{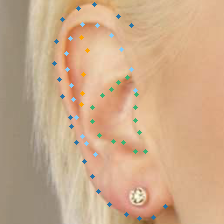}};
            
            \node (0_render) [below = 0.2cm of 0_lm, inner sep=0pt]
            {\includegraphics[width=4cm]{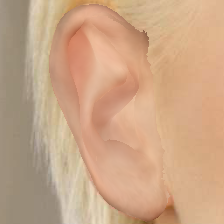}};
            
            \node (1_lm) [right = 0.2cm of 0_lm, inner sep=0pt]
            {\includegraphics[width=4cm]{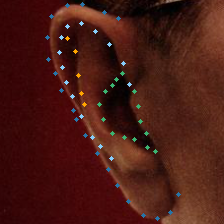}};
            
            \node (1_render) [below = 0.2cm of 1_lm, inner sep=0pt]
            {\includegraphics[width=4cm]{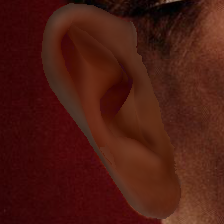}};
            
            \node (2_lm) [right = 0.2cm of 1_lm, inner sep=0pt]
            {\includegraphics[width=4cm]{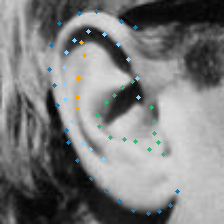}};
            
            \node (2_render) [below = 0.2cm of 2_lm, inner sep=0pt]
            {\includegraphics[width=4cm]{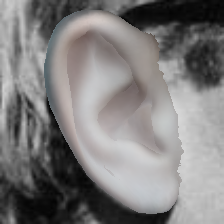}};
        \end{tikzpicture}
    }
    \caption{Test set prediction results with different ear colours. Top row: original ear images marked with predicted 2D landmarks. Bottom row: predicted 3D ear meshes projected onto original ear images.}
    \label{fig:qual-example-render-and-lm}
\end{figure}

\noindent Qualitative evaluations of this work focus on visually showing the 3D reconstruction results on ITWE-A's test set. In Figure \ref{fig:qual-example-render-and-lm}, three images with large colour variation are predicted, the top row shows the 2D landmark predictions look reasonable. The comparison between the top row and the bottom row shows that the quality of the reconstructed 3D meshes are reasonable in geometric aspect, while the in-the-wild colour model can reconstruct a large variation of in-the-wild ear colours even from grayscale images.

\begin{figure}
    \centering
    \resizebox{\columnwidth}{!}{
        \begin{tikzpicture}
            \node (small_pose_orig) [inner sep=0pt, anchor=south east] at (0,0)
            {\includegraphics[width=4cm]{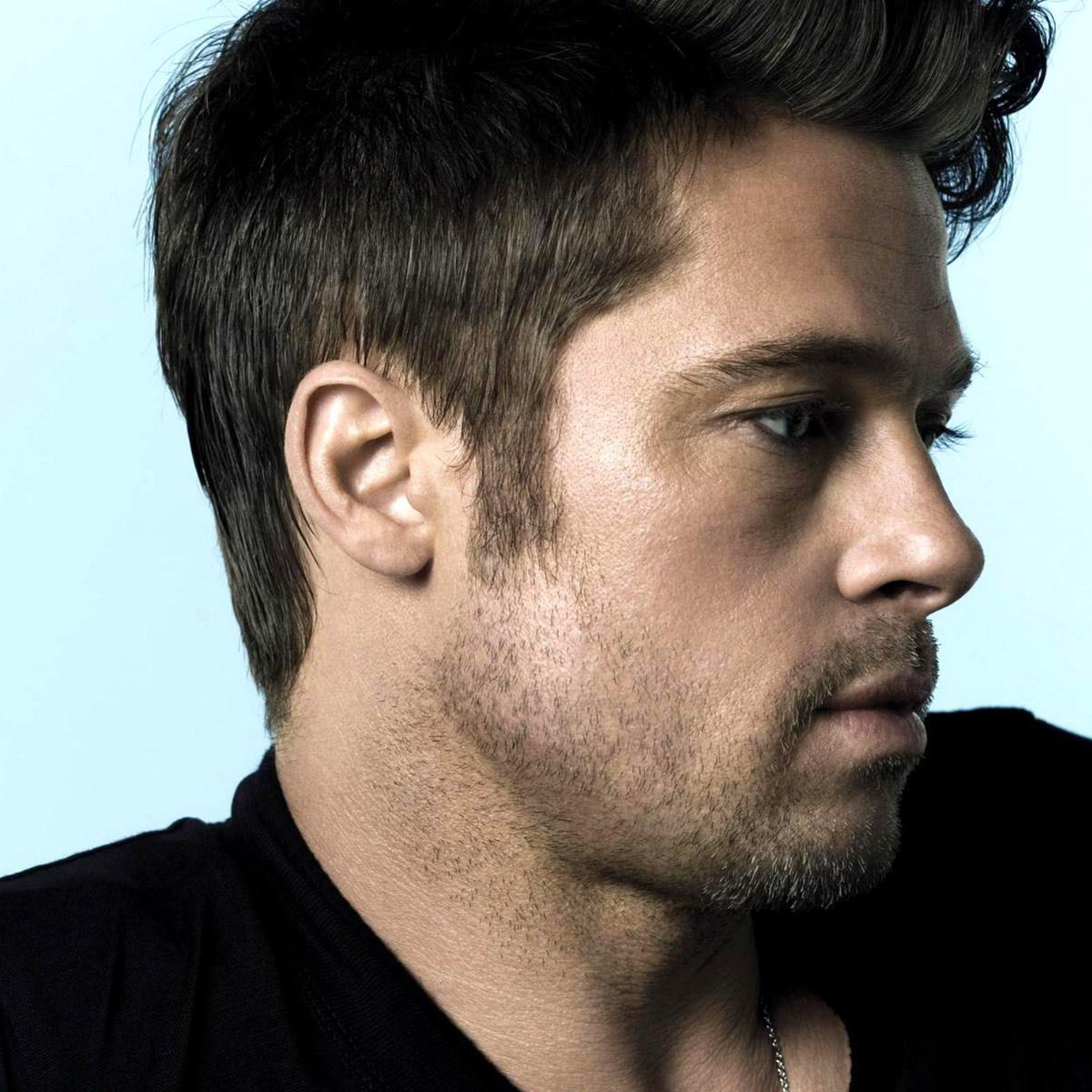}};
            
            \node (small_pose_mesh) [right = 0.2cm of small_pose_orig, inner sep=0pt]
            {\includegraphics[width=4cm]{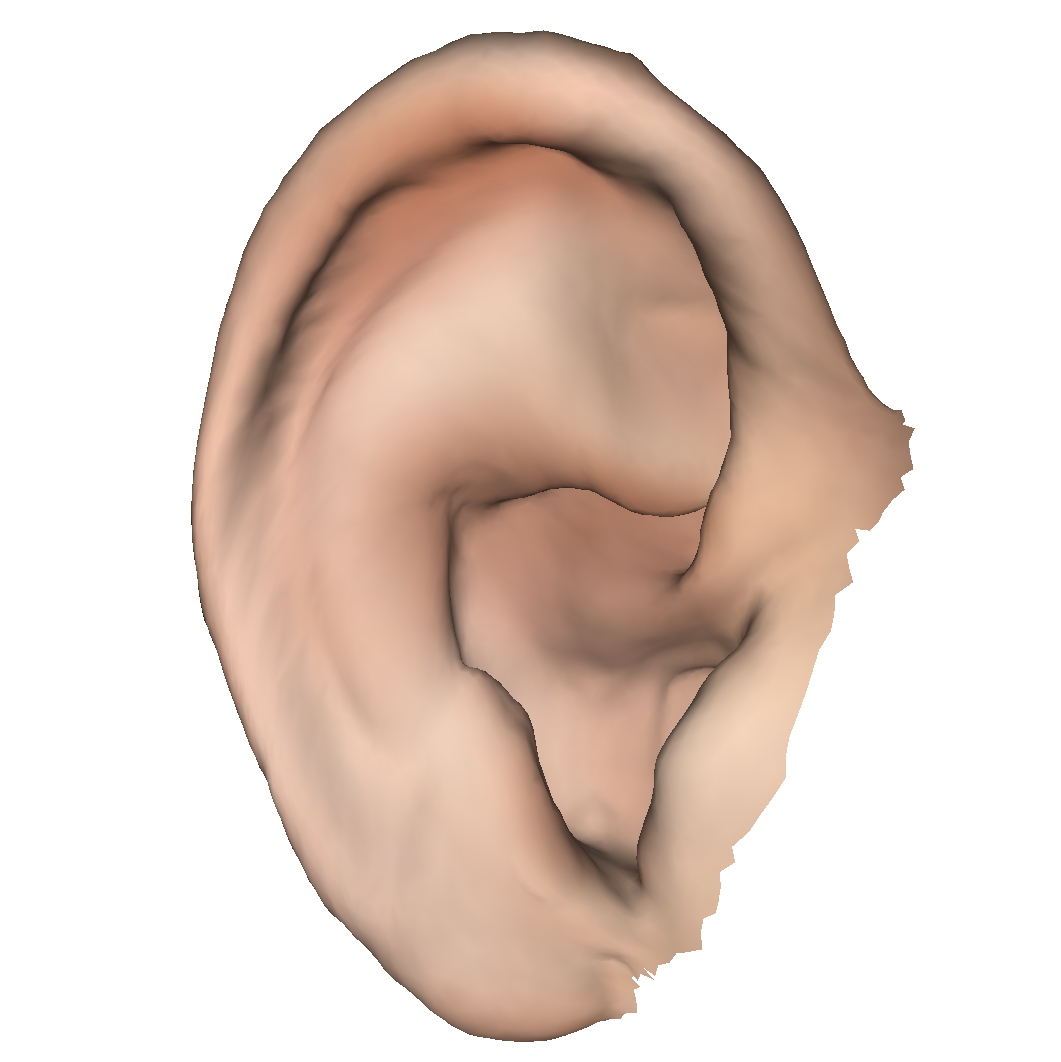}};
            
            \node (small_pose_lm) [right = 0.2cm of small_pose_mesh, inner sep=0pt]
            {\includegraphics[width=4cm]{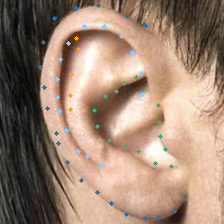}};
            
            \node (large_pose_orig) [below = 0.2cm of small_pose_orig, inner sep=0pt]
            {\includegraphics[width=4cm]{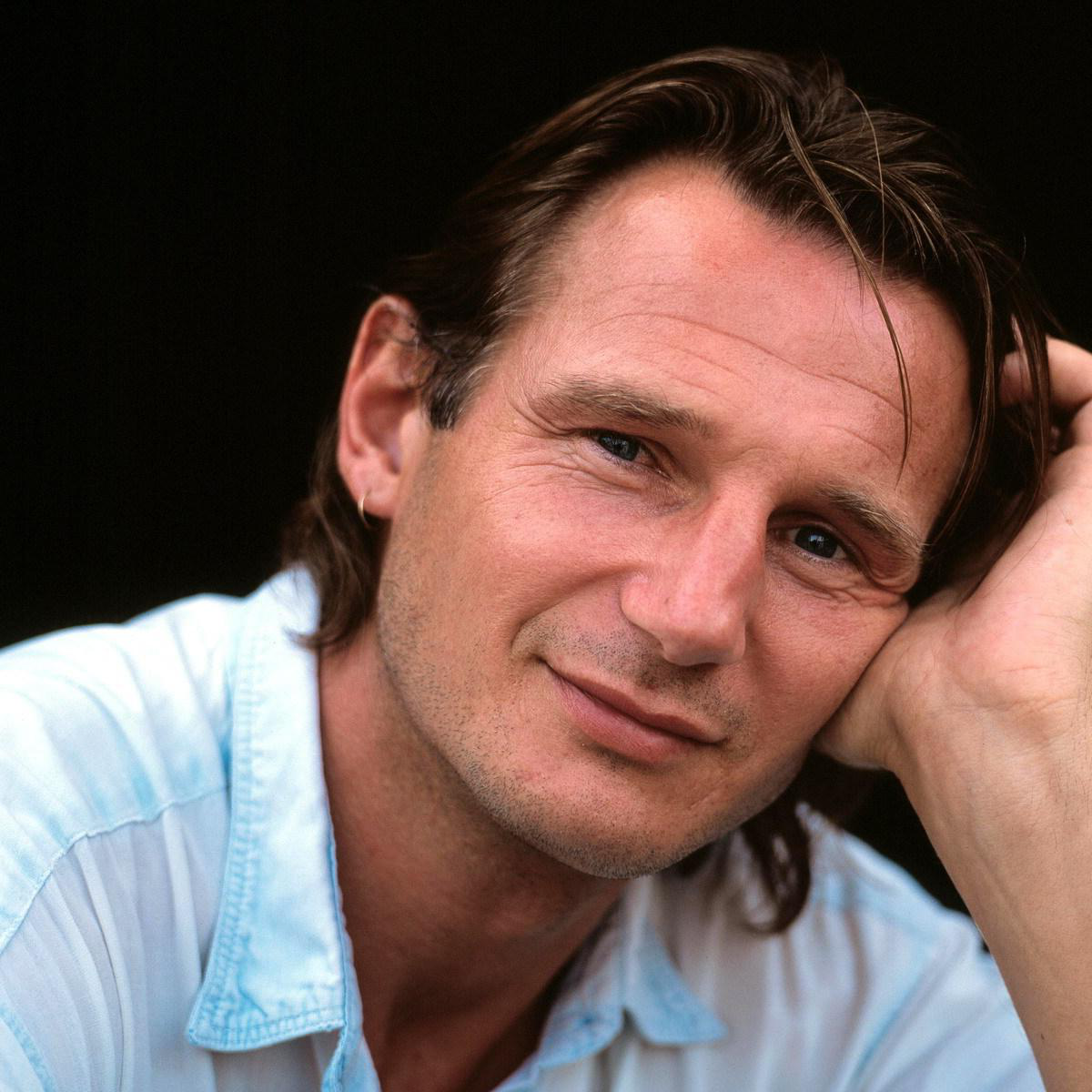}};
            
            \node (large_pose_mesh) [right = 0.2cm of large_pose_orig, inner sep=0pt]
            {\includegraphics[width=4cm]{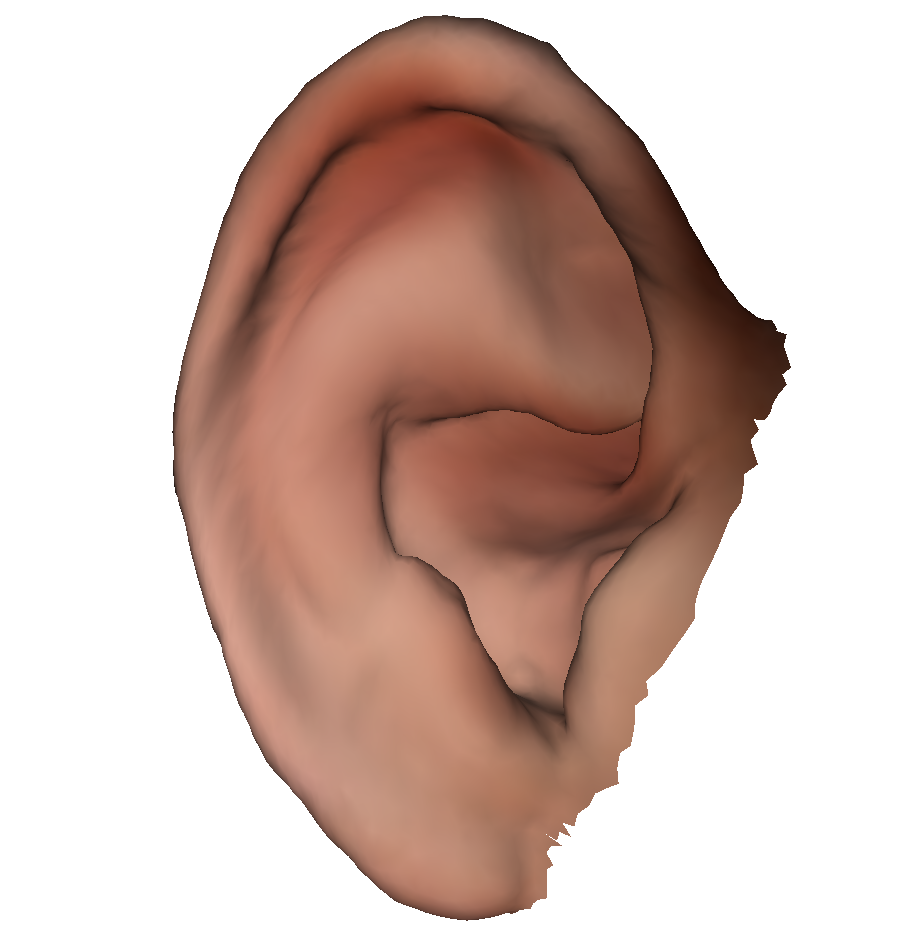}};
            
            \node (large_pose_lm) [right = 0.2cm of large_pose_mesh, inner sep=0pt]
            {\includegraphics[width=4cm]{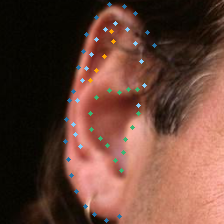}};

        \end{tikzpicture}
    }
    \caption{Test set prediction results with different head poses. Each row represents a distinct subject. \nth{1} column: Original uncropped images. \nth{2} column: Predicted 3D ear meshes. \nth{3} column: Predicted 2D landmarks. Ear pose is successfully predicted when difficult head pose involves.}
    \label{fig:qual-poses}
\end{figure}

In Figure \ref{fig:qual-poses}, two images with different head poses are selected for 3D ear reconstruction. The top row shows the results from a near-ideal head pose (\ie near-profile face) and the bottom row shows the results from a large head pose deviation from the ideal (\ie front facing, tilted head). The figure shows that HERA works well with different head poses. For the front facing images, the model predicts the correct horizontal rotation rather than narrowing the 3D ear mesh's width to match the 2D image.

\subsection{Ablation Study}

\begin{table}
    \caption{Normalised landmark distance error statistics on ITWE-A for ablation study.}
    \begin{center}
        \resizebox{\columnwidth}{!}{
            \begin{tabular}{|l|c|c|c|c|}
                \hline
                Method & mean $\pm$ std & median & $\leq 0.1$ & $\leq 0.06$\\
                \hline\hline
                HERA & $0.0398 \pm 0.009$ & $0.0391$ & $100\%$ & $96.2\%$ \\
                HERA-W/O-WTN & $0.0401 \pm 0.009$ & $0.0384$ & $100\%$ & $96.2\%$ \\
                HERA-W/O-PIX & $0.0392 \pm 0.009$ & $0.0387$ & $100\%$ & $96.2\%$ \\
                HERA-W/O-AUG & $0.0446 \pm 0.011$ & $0.0437$ & $100\%$ & $92.4\%$ \\
                HERA-W/O-AUG-LM & $0.0591 \pm 0.014$ & $0.0567$ & $99\%$ & $64.7\%$ \\
                \hline
            \end{tabular}
        }
    \end{center}
    
    \label{tab:abla-res}
\end{table}

\begin{figure}
    \centering
    \resizebox{\columnwidth}{!}{
        \begin{tikzpicture}
            \node (ablation_3dera) [inner sep=0pt, anchor=south east, label=below:(1)] at (0,0)
            {\includegraphics[width=4cm]{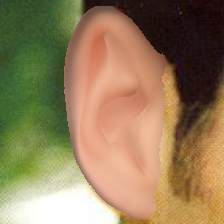}};
            
            \node (ablation_wo_pix) [right = 0.2cm of ablation_3dera, inner sep=0pt, label=below:(2)]
            {\includegraphics[width=4cm]{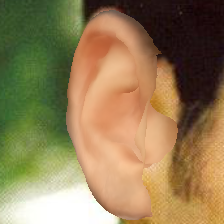}};
            
            \node (ablation_wo_whitening) [right = 0.2cm of ablation_wo_pix, inner sep=0pt, label=below:(3)]
            {\includegraphics[width=4cm]{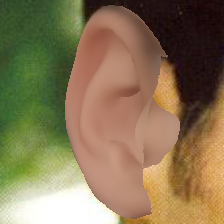}};
        \end{tikzpicture}
    }
    \caption{Appearance comparison between the reconstructed 3D ear meshes of (1) HERA, (2) HERA-W/O-WTN and (3) HERA-W/O-PIX. Only HERA reconstructs the external auditory canal part correctly.}
    \label{fig:abla-wtn-pix}
\end{figure}

\begin{figure}
    \centering
    \resizebox{\columnwidth}{!}{
        \begin{tikzpicture}
            \node (ablation_3dera_2) [inner sep=0pt, anchor=south east, label=below:(1)] at (0,0)
            {\includegraphics[width=4cm]{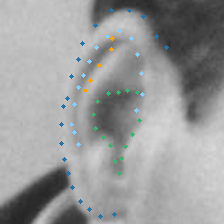}};
            
            \node (ablation_wo_aug) [right = 0.2cm of ablation_3dera_2, inner sep=0pt, label=below:(2)]
            {\includegraphics[width=4cm]{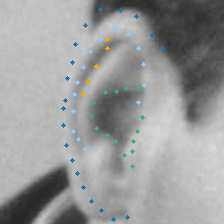}};
            
            \node (ablation_wo_aug_lm) [right = 0.2cm of ablation_wo_aug, inner sep=0pt, label=below:(3)]
            {\includegraphics[width=4cm]{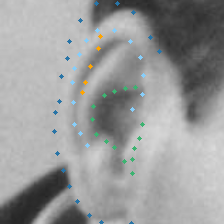}};
        \end{tikzpicture}
    }
    \caption{2D landmark localisation comparison between the prediction results of (1) HERA, (2) HERA-W/O-AUG and (3) HERA-W/O-AUG-LM. Data augmentation enables better ear rotation prediction and landmark loss is vital to accurate alignment especially for the ear contour part.}
    \label{fig:abla-aug-lm}
\end{figure}

\noindent We now study how each component can affect HERA's performance and we evaluate on several system variations including HERA-W/O-WTN (without PCA whitening on 3D ear shape parameters $\bm{\beta}_{s}$), HERA-W/O-PIX (without pixel loss), HERA-W/O-AUG (without data augmentation) and HERA-W/O-AUG-LM (without landmark loss). Table \ref{tab:abla-res} shows the statistics for all the variations of HERA. When training without PCA whitening on 3D ear shape parameters and without pixel loss, their performances on 2D landmark localisation are similar to the final proposed method. However, as shown in Figure \ref{fig:abla-wtn-pix}, both variations failed to predict the projection of the external auditory canal part of the ear correctly. An imbalanced intermediate code vector is one of the potential reasons why the variation without PCA whitening on shape parameters fails on the external auditory canal part. In any case, a balanced design of intermediate code vector with similar variance for each parameter can benefit the performance of the neural network. The version without pixel loss focuses on lowering the landmark alignment error regardless of the overall appearance of the ear. Therefore it is necessary to utilise the pixel loss.

When training without data augmentation, the 2D landmark localisation performance drops by a small amount mainly due to its lack of variety in ear rotation, shown in Figure \ref{fig:abla-aug-lm}. When training without landmark loss, the predicted landmarks is not accurate enough, shown in Figure \ref{fig:abla-aug-lm}. As a result, the reconstructed 3D ears are not accurately aligned with the 2D ears especially for the ear contours.

\section{\uppercase{Conclusion}}

\noindent As a large proportion of human-related 3D reconstruction approaches focus on the human face, 3D ear reconstruction, as an important human-related task, has much less related work. 
In this paper, we propose a self-supervised deep 3D ear reconstruction autoencoder from single image. Our model reconstructs the 3D ear mesh with a plausible appearance and accurate dense alignment, as witnessed by the accurate alignment compared to ground truth landmarks. The comprehensive evaluation shows that our method achieves state-of-the-art performance in 3D ear reconstruction and 3D ear alignment.

%Inspired by MoFA's elegant autoencoder structure for self-supervised 3D face reconstruction, we have employed such a structure on the 3D ear reconstruction task by employing the York Ear Model, an in-the-wild ear colour model and PCA whitening.

%% Body text ends.

\bibliographystyle{apalike}
{\small
\bibliography{main}}

\begin{thebibliography}{}

\bibitem[Bizjak et~al., 2019]{bizjak2019mask}
Bizjak, M., Peer, P., and Emer{\v{s}}i{\v{c}}, {\v{Z}}. (2019).
\newblock Mask r-cnn for ear detection.
\newblock In {\em 2019 42nd International Convention on Information and
  Communication Technology, Electronics and Microelectronics (MIPRO)}, pages
  1624--1628. IEEE.

\bibitem[Blanz and Vetter, 1999]{blanz1999morphable}
Blanz, V. and Vetter, T. (1999).
\newblock A morphable model for the synthesis of 3d faces.
\newblock In {\em Proceedings of the 26th annual conference on Computer
  graphics and interactive techniques}, pages 187--194.

\bibitem[Cootes et~al., 1998]{cootes1998active}
Cootes, T.~F., Edwards, G.~J., and Taylor, C.~J. (1998).
\newblock Active appearance models.
\newblock In {\em European conference on computer vision}, pages 484--498.
  Springer.

\bibitem[Dai et~al., 2020]{dai20203d}
Dai, H., Pears, N., Huber, P., and Smith, W.~A. (2020).
\newblock 3d morphable models: The face, ear and head.
\newblock In {\em 3D Imaging, Analysis and Applications}, pages 463--512.
  Springer.

\bibitem[Dai et~al., 2018]{dai2018data}
Dai, H., Pears, N., and Smith, W. (2018).
\newblock A data-augmented 3d morphable model of the ear.
\newblock In {\em 2018 13th IEEE International Conference on Automatic Face \&
  Gesture Recognition (FG 2018)}, pages 404--408. IEEE.

\bibitem[Deng et~al., 2009]{deng2009imagenet}
Deng, J., Dong, W., Socher, R., Li, L.-J., Li, K., and Fei-Fei, L. (2009).
\newblock Imagenet: A large-scale hierarchical image database.
\newblock In {\em 2009 IEEE conference on computer vision and pattern
  recognition}, pages 248--255. Ieee.

\bibitem[Emer{\v{s}}i{\v{c}} et~al., 2017a]{emervsivc2017pixel}
Emer{\v{s}}i{\v{c}}, {\v{Z}}., Gabriel, L.~L., {\v{S}}truc, V., and Peer, P.
  (2017a).
\newblock Pixel-wise ear detection with convolutional encoder-decoder networks.
\newblock {\em arXiv preprint arXiv:1702.00307}.

\bibitem[Emer{\v{s}}i{\v{c}} et~al., 2017b]{emervsivc2017ear}
Emer{\v{s}}i{\v{c}}, {\v{Z}}., {\v{S}}truc, V., and Peer, P. (2017b).
\newblock Ear recognition: More than a survey.
\newblock {\em Neurocomputing}, 255:26--39.

\bibitem[Emer{\v{s}}i{\v{c}} et~al., 2019]{emervsivc2019unconstrained}
Emer{\v{s}}i{\v{c}}, {\v{Z}}., SV, A.~K., Harish, B., Gutfeter, W., Khiarak,
  J., Pacut, A., Hansley, E., Segundo, M.~P., Sarkar, S., Park, H., et~al.
  (2019).
\newblock The unconstrained ear recognition challenge 2019.
\newblock In {\em 2019 International Conference on Biometrics (ICB)}, pages
  1--15. IEEE.

\bibitem[Gecer et~al., 2019]{gecer2019ganfit}
Gecer, B., Ploumpis, S., Kotsia, I., and Zafeiriou, S. (2019).
\newblock Ganfit: Generative adversarial network fitting for high fidelity 3d
  face reconstruction.
\newblock In {\em Proceedings of the IEEE Conference on Computer Vision and
  Pattern Recognition}, pages 1155--1164.

\bibitem[Hansley et~al., 2018]{hansley2018employing}
Hansley, E.~E., Segundo, M.~P., and Sarkar, S. (2018).
\newblock Employing fusion of learned and handcrafted features for
  unconstrained ear recognition.
\newblock {\em IET Biometrics}, 7(3):215--223.

\bibitem[He et~al., 2016]{he2016deep}
He, K., Zhang, X., Ren, S., and Sun, J. (2016).
\newblock Deep residual learning for image recognition.
\newblock In {\em Proceedings of the IEEE conference on computer vision and
  pattern recognition}, pages 770--778.

\bibitem[Kessy et~al., 2018]{kessy2018optimal}
Kessy, A., Lewin, A., and Strimmer, K. (2018).
\newblock Optimal whitening and decorrelation.
\newblock {\em The American Statistician}, 72(4):309--314.

\bibitem[Liu et~al., 2016]{liu2016joint}
Liu, F., Zeng, D., Zhao, Q., and Liu, X. (2016).
\newblock Joint face alignment and 3d face reconstruction.
\newblock In {\em European Conference on Computer Vision}, pages 545--560.
  Springer.

\bibitem[McDonagh and Tzimiropoulos, 2016]{mcdonagh2016joint}
McDonagh, J. and Tzimiropoulos, G. (2016).
\newblock Joint face detection and alignment with a deformable hough transform
  model.
\newblock In {\em European Conference on Computer Vision}, pages 569--580.
  Springer.

\bibitem[Paszke et~al., 2019]{paszke2019pytorch}
Paszke, A., Gross, S., Massa, F., Lerer, A., Bradbury, J., Chanan, G., Killeen,
  T., Lin, Z., Gimelshein, N., Antiga, L., et~al. (2019).
\newblock Pytorch: An imperative style, high-performance deep learning library.
\newblock In {\em Advances in neural information processing systems}, pages
  8026--8037.

\bibitem[Ploumpis et~al., 2020]{ploumpis2020towards}
Ploumpis, S., Ververas, E., O'Sullivan, E., Moschoglou, S., Wang, H., Pears,
  N., Smith, W., Gecer, B., and Zafeiriou, S.~P. (2020).
\newblock Towards a complete 3d morphable model of the human head.
\newblock {\em IEEE Transactions on Pattern Analysis and Machine Intelligence}.

\bibitem[Ravi et~al., 2020]{ravi2020pytorch3d}
Ravi, N., Reizenstein, J., Novotny, D., Gordon, T., Lo, W.-Y., Johnson, J., and
  Gkioxari, G. (2020).
\newblock Accelerating 3d deep learning with pytorch3d.
\newblock {\em arXiv:2007.08501}.

\bibitem[Richardson et~al., 2016]{richardson20163d}
Richardson, E., Sela, M., and Kimmel, R. (2016).
\newblock 3d face reconstruction by learning from synthetic data.
\newblock In {\em 2016 fourth international conference on 3D vision (3DV)},
  pages 460--469. IEEE.

\bibitem[Tewari et~al., 2017]{tewari2017mofa}
Tewari, A., Zollhofer, M., Kim, H., Garrido, P., Bernard, F., Perez, P., and
  Theobalt, C. (2017).
\newblock Mofa: Model-based deep convolutional face autoencoder for
  unsupervised monocular reconstruction.
\newblock In {\em Proceedings of the IEEE International Conference on Computer
  Vision Workshops}, pages 1274--1283.

\bibitem[Tran and Liu, 2018]{tran2018nonlinear}
Tran, L. and Liu, X. (2018).
\newblock Nonlinear 3d face morphable model.
\newblock In {\em Proceedings of the IEEE conference on computer vision and
  pattern recognition}, pages 7346--7355.

\bibitem[Zhou and Zaferiou, 2017]{zhou2017deformable}
Zhou, Y. and Zaferiou, S. (2017).
\newblock Deformable models of ears in-the-wild for alignment and recognition.
\newblock In {\em 2017 12th IEEE International Conference on Automatic Face \&
  Gesture Recognition (FG 2017)}, pages 626--633. IEEE.

\bibitem[Zhu et~al., 2017]{zhu2017face}
Zhu, X., Liu, X., Lei, Z., and Li, S.~Z. (2017).
\newblock Face alignment in full pose range: A 3d total solution.
\newblock {\em IEEE transactions on pattern analysis and machine intelligence},
  41(1):78--92.

\end{thebibliography}

\end{document}